\documentclass[11pt]{article}

\usepackage[]{EMNLP2022}

\usepackage{times}
\usepackage{latexsym}

\usepackage[T1]{fontenc}

\usepackage[utf8]{inputenc}

\usepackage[ruled,linesnumbered,vlined]{algorithm2e}
\SetAlFnt{\small}
\SetAlCapFnt{\small}
\SetAlCapNameFnt{\small}
\newcommand{\var}{\texttt}
\let\oldnl\nl
\newcommand{\nonl}{\renewcommand{\nl}{\let\nl\oldnl}}

\usepackage{amsfonts,amssymb}
\usepackage{bbm}
\usepackage{multirow}
\usepackage{amsmath}
\usepackage{booktabs} 
\usepackage{tablefootnote}
\usepackage{graphicx}
\usepackage{caption}
\usepackage{subcaption}
\usepackage{makecell}
\usepackage{bbding}
\usepackage{color}
\usepackage{arydshln} 

\newcommand{\fedr}{\textsc{FedR}}
\newcommand{\fede}{\textsc{FedE}}

\usepackage{microtype}

\usepackage{inconsolata}

\title{Efficient Federated Learning on Knowledge Graphs via \\ Privacy-preserving Relation Embedding Aggregation} 

\author{Kai Zhang\textsuperscript{1}, 
Yu Wang\textsuperscript{2}, Hongyi Wang\textsuperscript{3}, Lifu Huang\textsuperscript{4}, Carl Yang\textsuperscript{5}, Xun Chen\textsuperscript{6}, Lichao Sun\textsuperscript{1} \\
\textsuperscript{1}Lehigh University, \textsuperscript{2}University of Illinois Chicago, 
\textsuperscript{3}Carnegie Mellon University,\\ \textsuperscript{4}Virginia Tech, \textsuperscript{5}Emory University,
\textsuperscript{6}Samsung Research America \\
\texttt{kaz321@lehigh.edu, ywang617@uic.edu, hongyiwa@andrew.cmu.edu,} \\
\texttt{lifuh@vt.edu, j.carlyang@emory.edu, xun.chen@samsung.com, lis221@lehigh.edu}
}

\begin{document}
\maketitle
\begin{abstract}
Federated learning (FL) can be essential in knowledge representation, reasoning, and data mining applications over multi-source knowledge graphs (KGs). A recent study FedE first proposes an FL framework that shares entity embeddings of KGs across all clients. However, entity embedding sharing from FedE would incur a severe privacy leakage. Specifically, the known entity embedding can be used to infer whether a specific relation between two entities exists in a private client. In this paper, we introduce a novel attack method that aims to recover the original data based on the embedding information, which is further used to evaluate the vulnerabilities of FedE. Furthermore, we propose a \textbf{Fed}erated learning paradigm with privacy-preserving \textbf{R}elation embedding aggregation (\fedr) to tackle the privacy issue in FedE. Besides, relation embedding sharing can significantly reduce the communication cost due to its smaller size of queries. We conduct extensive experiments to evaluate \fedr{} with five different KG embedding models and three datasets. Compared to FedE, \fedr{} achieves similar utility and significant improvements regarding privacy-preserving effect and communication efficiency on the link prediction task.
\end{abstract}

\section{Introduction}

Knowledge graphs (KGs) are critical data structures to represent human knowledge and serve as resources for various real-world applications, such as recommendation and question answering \cite{gong2021smr, liu2018t}. However, most KGs are usually incomplete and naturally distributed to different clients. Despite each client can explore the missing links with their own KGs by knowledge graph embedding (KGE) models \citep{lin2015learning}, exchanging knowledge with others can further enhance completion performance because the overlapping elements are usually involved in different KGs \citep{chen2021fede, peng2021differentially}.

\begin{figure}
    \centering
    \includegraphics[width=0.48\textwidth]{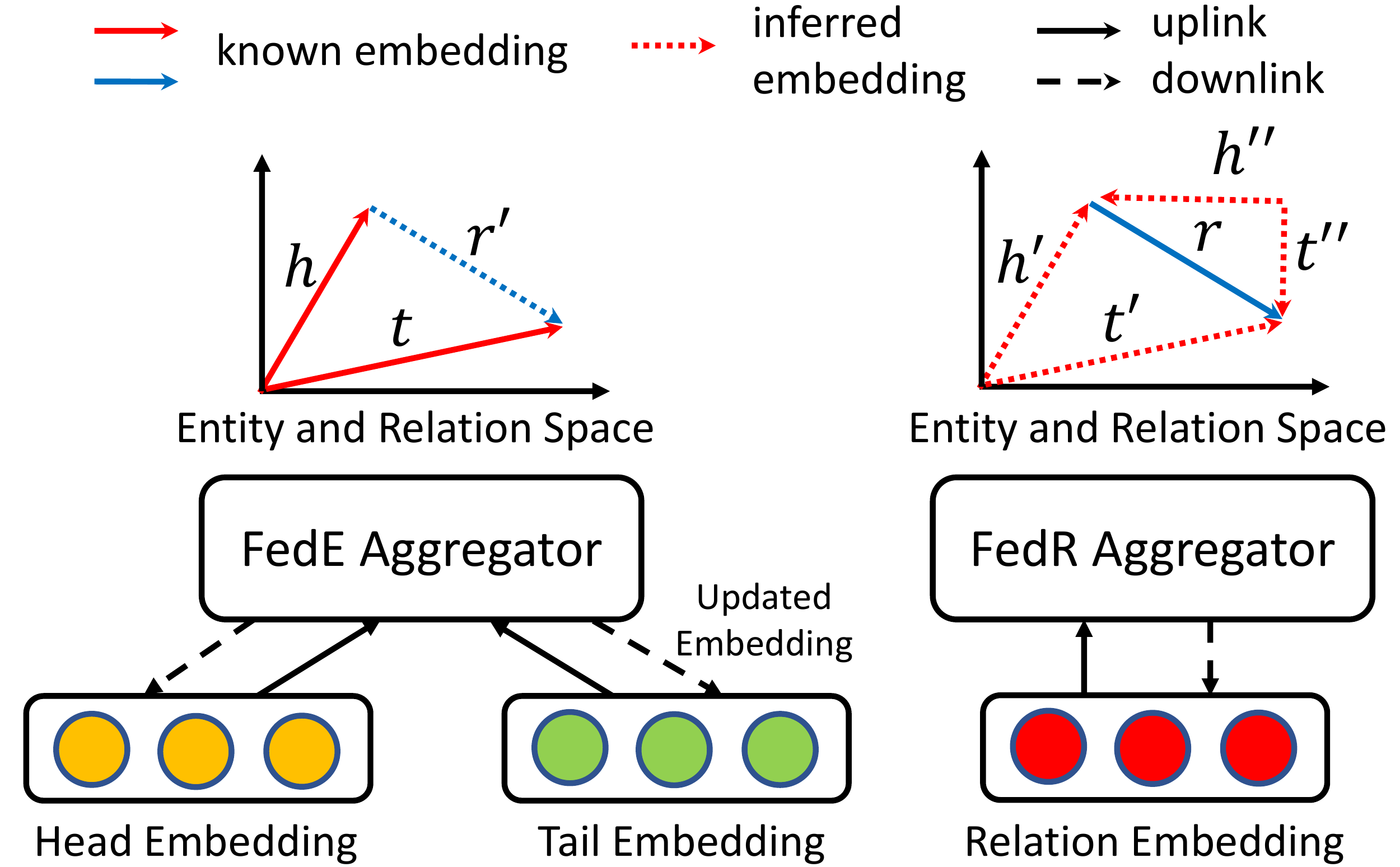}
    \caption{FedE aggregates entity embeddings from clients while \fedr{} aggregates relation embeddings. Since in \fedr{}, there would be infinite embedding pairs of head and tail given a relation embedding, the inference attack would fail.}
    \vspace{-0.5cm}
    \label{fig:overview}
\end{figure}

To exchange knowledge, the first federated learning (FL) framework for KG -- FedE is recently proposed, where each client trains local embeddings on its KG while the server receives and aggregates only locally-computed updates of entity embeddings instead of collecting triplets directly ~\citep{chen2021fede}. However, at the very beginning in FedE, the server should collect the entity sets of every client for entity alignment, which will lead to unintentional privacy leakage: 1) entity's information, such as the customer's name, is usually sensitive but it is fully exposed to the server; 2) the relation embedding will be inferred and be exploited for knowledge graph reconstruction attack if there exists the malicious server (see Section \ref{sec:privacy_intro}). Therefore, we propose \fedr{} that adopts relation embedding aggregation to tackle the privacy issue in FedE. The major difference is shown in Figure \ref{fig:overview}. Besides, the number of entities is usually greater than the number of relations in real-world graph databases, so sharing relation embedding is more communication-efficient.

We summarize the following contributions of our work. 1) We present a KG reconstruction attack method and reveal that FedE suffers a potential privacy leakage due to a malicious server and its colluded clients. 2) We propose \fedr{}, an efficient and privacy-preserving FL framework on KGs. Experimental results demonstrate that \fedr{} has the competitive performance compared with FedE, but gains substantial improvements in terms of privacy-preserving effect and communication efficiency.


\section{Background} \label{sec:back}
\paragraph{Knowledge graph and its embedding.} KG is a directed multi-relational graph whose nodes correspond to entities and edges of the form (head, relation, tail), which is denoted as a triplet $(h,r,t)$. KGE model aims to learn low-dimensional representations of elements in a KG via maximizing scoring function $f(\mathbf{h,r,t})$ of all embedding of triplets. In other words, as depicted in Figure \ref{fig:overview}, we can infer relation embedding in terms of $\mathbf{r'}=\arg\max_{\mathbf{r}} f(\mathbf{h,r,t})$ given entity embeddings, but we cannot obtain $\mathbf{t'}=\arg\max_{\mathbf{t}} f(\mathbf{h,r,t})$ merely based on known relation embedding $\mathbf{r}$. 

\paragraph{Federated learning and FedE.} FL allows different clients to collaboratively learn a global model without sharing their local data \citep{mcmahan2017communication}. In particular, the aim is to minimize: $\min _{w} f(w)=\mathbb{E}_{k}\left[F_{k}(w)\right]$, where $F_{k}(w)$ is the local objective that measures the local empirical risk of $k$-th client. Compared to model sharing in vanilla FL 
, FedE introduces a new mechanism that aggregates only entity embedding. More concretely, the server maintains a complete table including entity embeddings and the corresponding entity IDs, and the server can identify if an entity exists in a client for entity alignment. 





\section{Methodology} \label{sec:method}

\subsection{Knowledge Graph Reconstruction}
\label{sec:privacy_intro}

The purpose of knowledge graph reconstruction attack is to recover original entities and relations in a KG given traitor's information including parital or all triplets and the corresponding embeddings, namely element-embedding pairs. The attack procedure for FedE is summarized as follows (suppose there is a malicious server and one traitor):

\textbf{1)} The server colludes with one client C1 to obtain its element-embedding pairs $\langle (E,\mathbf{e}), (R,\mathbf{r}) \rangle$.\\
\indent \textbf{2)} Infer the target client's relation embedding by calculating $\mathbf{r'}=\arg\max_{\mathbf{r}} f(\mathbf{h,r,t})$.\\
\indent \textbf{3)} Measure the discrepancy between the inferred element embedding such as relation embedding $\mathbf{r'}$ and all known $\mathbf{r}$ with cosine similarity.\\
\indent \textbf{4)} Infer the relation $R'$ as $R$, $E'$ as $E$ with corresponding largest similarity scores. Then the target client's KG/triplet can be reconstructed. More detials are included in Appendix \ref{sec:kg_attack}.



\textbf{Privacy leakage quantization in FedE.} We define two metrics: \textit{Triplet Reconstruction Rate} (TRR) and \textit{Entity Reconstruction Rate} (ERR) to measure the ratio of corretly reconstructed triplets and entities to the relevant whole number of elements, respectively. 
We let the server owns 30\%, 50\%, 100\% trained element-embedding pairs from C1, the traitor, to reconstruct entities and triplets of others. 
The results of privacy leakage on FB15k-237 \cite{toutanova2015representing} over three clients are summarized in Table \ref{tab:privacy_fb15k}. LR in the table denotes information (element-embedding pairs) leakage ratio from C1. It is clear that the server only needs to collude with one client to obtain most of the information of KGs on other clients. In a word, FedE is not privacy-preserving. 

\begin{table}[]
\centering
\setlength{\tabcolsep}{3.8pt}
\small
\begin{tabular}{lcccccc}
\toprule
\multirow{2}{*}{LR} & \multicolumn{2}{c}{30\%} & \multicolumn{2}{c}{50\%} & \multicolumn{2}{c}{100\%} \\ \cmidrule{2-7} 
         & ERR    & TRR    & ERR    & TRR    & ERR    & TRR    \\ \midrule
C2 & 0.2904 & 0.0607 & 0.4835 & 0.1951 & 0.9690 & 0.7378 \\
C3 & 0.2906 & 0.0616 & 0.4846 & 0.1956 & 0.9685 & 0.7390 \\ \bottomrule
\end{tabular}
\caption{Privacy leakage on FB15k-237 with TransE.}
\label{tab:privacy_fb15k}
\vspace{-10pt}
\end{table}

\begin{table*}[t]
\centering
\setlength{\tabcolsep}{3.4pt}
\small
\begin{tabular}{cccccccccccccc}
\toprule
\multicolumn{2}{c|}{Dataset} & \multicolumn{4}{c|}{DDB14} & \multicolumn{4}{c|}{WN18RR} & \multicolumn{4}{c}{FB15k-237} \\ \hline
\multicolumn{1}{c|}{Model} & \multicolumn{1}{c|}{Setting} & C = 5 & C = 10 & C = 15 & \multicolumn{1}{c|}{C = 20} & C = 5 & C = 10 & C = 15 & \multicolumn{1}{c|}{C = 20} & C = 5 & C = 10 & C = 15 & C = 20 \\ \hline

\multicolumn{1}{c|}{\multirow{3}{*}{TransE}} & 
\multicolumn{1}{c|}{\var{Local}} &0.4206  &0.2998  &0.2464  & \multicolumn{1}{c|}{0.2043} &0.0655  &0.0319  &0.0378  & \multicolumn{1}{c|}{0.0285} &0.2174  &0.1255  &0.1087  &0.0874  \\
\multicolumn{1}{c|}{} & 
\multicolumn{1}{c|}{FedE} & 0.4572 & 0.3493 & 0.3076 & \multicolumn{1}{c|}{0.2962} & 0.1359 & 0.1263 & 0.1204 & \multicolumn{1}{c|}{0.1419} & 0.2588 & 0.2230 & 0.2065 & 0.1892 \\
\multicolumn{1}{c|}{} & 
\multicolumn{1}{c|}{\fedr{}} & \textbf{\underline{0.4461}} & \underline{0.3289} & \underline{0.2842} & \multicolumn{1}{c|}{\underline{0.2761}} & \underline{0.0859} & \underline{0.0779} & \underline{0.0722} & \multicolumn{1}{c|}{\underline{0.0668}} & \textbf{\underline{0.2520}} & \underline{0.2052} & \underline{0.1867} & \underline{0.1701} \\ \hline

\multicolumn{1}{c|}{\multirow{3}{*}{RotatE}} & 
\multicolumn{1}{c|}{\var{Local}} &0.4187  &0.2842  &0.2411  & \multicolumn{1}{c|}{0.2020} &0.1201  &0.0649  &0.0513  & \multicolumn{1}{c|}{0.0155} &0.2424  &0.1991  &0.1526  &0.0860  \\
\multicolumn{1}{c|}{} & 
\multicolumn{1}{c|}{FedE} & 0.4667 & 0.3635 & 0.3244 & \multicolumn{1}{c|}{0.3031} & 0.2741 & 0.1936 & 0.1287 & \multicolumn{1}{c|}{0.0902} & 0.2682 & 0.2278 & 0.2199 & 0.1827 \\
\multicolumn{1}{c|}{} & 
\multicolumn{1}{c|}{\fedr{}} & \underline{0.4477} & \underline{0.3184} & \underline{0.2765} & \multicolumn{1}{c|}{\underline{0.2681}} & \underline{0.1372} & \underline{0.1271} & \underline{0.1074} & \multicolumn{1}{c|}{\textbf{\underline{0.0912}}} & \underline{0.2510} & \underline{0.2080} & \underline{0.1854} & \underline{0.1586} \\ \hline

\multicolumn{1}{c|}{\multirow{3}{*}{DistMult}} & \multicolumn{1}{c|}{\var{Local}} &0.2248  &0.1145  &0.0764  & \multicolumn{1}{c|}{0.0652} &0.0654  &0.0517  &0.0548  & \multicolumn{1}{c|}{0.0374} &0.1133  &0.0773  &0.0765 &0.0689  \\
\multicolumn{1}{c|}{} & \multicolumn{1}{c|}{FedE} & 0.3037 & 0.2485 & 0.2315 & \multicolumn{1}{c|}{0.1877} & 0.1137 & 0.0946 & 0.0766 & \multicolumn{1}{c|}{0.0670} & 0.1718 & 0.1129 & 0.0901 & 0.0753 \\
\multicolumn{1}{c|}{} & \multicolumn{1}{c|}{\fedr{}} & \textbf{\underline{0.4219}} & \textbf{\underline{0.3146}} & \textbf{\underline{0.2685}} & \multicolumn{1}{c|}{\textbf{\underline{0.2577}}} & \textbf{\underline{0.1350}} & \textbf{\underline{0.1202}} & \textbf{\underline{0.1198}} & \multicolumn{1}{c|}{\textbf{\underline{0.0898}}} & \textbf{\underline{0.1670}} & \underline{0.0999} & \textbf{\underline{0.0884}} & \textbf{\underline{0.0814}} \\ \hline

\multicolumn{1}{c|}{\multirow{3}{*}{ComplEx}} & \multicolumn{1}{c|}{\var{Local}} &0.3406  &0.2025  &0.1506  & \multicolumn{1}{c|}{0.1247} &0.0035  &0.0033  &0.0033  & \multicolumn{1}{c|}{0.0022} &0.1241  &0.0694  &0.0571  &0.0541  \\
\multicolumn{1}{c|}{} & \multicolumn{1}{c|}{FedE} & 0.3595 & 0.2838 & 0.2411 & \multicolumn{1}{c|}{0.1946} & 0.0153 & 0.0115 & 0.0108 & \multicolumn{1}{c|}{0.0122} & 0.1603 & 0.1161 & 0.0944 & 0.0751 \\
\multicolumn{1}{c|}{} & \multicolumn{1}{c|}{\fedr{}} & \textbf{\underline{0.4287}} & \textbf{\underline{0.3235}} & \textbf{\underline{0.2747}} & \multicolumn{1}{c|}{\textbf{\underline{0.2611}}} & \textbf{\underline{0.0203}} & \textbf{\underline{0.0152}} & \textbf{\underline{0.0152}} & \multicolumn{1}{c|}{\textbf{\underline{0.0166}}} & \textbf{\underline{0.1716}} & \textbf{\underline{0.1174}}& \textbf{\underline{0.1075}} & \textbf{\underline{0.0993}} \\ \hline

\multicolumn{1}{c|}{\multirow{3}{*}{NoGE}} & \multicolumn{1}{c|}{\var{Local}} &0.3178  &0.2298  &0.1822  & \multicolumn{1}{c|}{0.1580} &0.0534  &0.0474  &0.0371  & \multicolumn{1}{c|}{0.0372} &0.2315  &0.1642  &0.1246  &0.1042  \\
\multicolumn{1}{c|}{} & \multicolumn{1}{c|}{FedE} &0.3193  &0.3171 &0.2678  & \multicolumn{1}{c|}{0.2659} &0.0789 &0.0697  &0.0632  & \multicolumn{1}{c|}{0.0533} &0.2412  &0.1954 &0.1730  &0.1637  \\
\multicolumn{1}{c|}{} & \multicolumn{1}{c|}{\fedr{}} &\textbf{\underline{0.4312}}  &\textbf{\underline{0.3127}}  &\textbf{\underline{0.2604}}  & \multicolumn{1}{c|}{\underline{0.2452}} &\underline{0.0669}  &\underline{0.0543} &\underline{0.0530}  & \multicolumn{1}{c|}{\underline{0.0499}} &\textbf{\underline{0.2432}} &\underline{0.1822}  &\underline{0.1448}  &\underline{0.1282}  \\ \bottomrule
\end{tabular}
\vspace{-0.2cm}
\caption{Link prediction results (MRR). \textbf{Bold} number denotes \fedr{} performs better than or close to (within 3\% performance decrease) FedE. \underline{Underline} number denotes the better result between \fedr{} and \var{Local}.}
\vspace{-10pt}
\label{tab:effect}
\end{table*}

\begin{algorithm}
  \SetCommentSty{mycommfont}
  \SetKwInOut{Input}{Input}
  \SetKwInOut{Output}{output}
  \Input{local datasets $T^{c}$, number of clients $C$, number of local epochs $E$, learning rate $\eta$}
  \BlankLine
  \nonl \textbf{Server excutes:}\\
  collect relations from clients via \var{PSU}\\
  initialize relation table with relation embedding $\mathbf{E}_{0}^r$ \\
  \For{\textup{round} $t = 0,1,...$}{
    \textup{Send the relation table to all clients}\\
    \textup{Sample a set of clients} $C_t$\\
    \ForPar{$c \in C_t$}{
        $\mathbf{E}_{t+1}^{r,c}, \mathbf{v}^c \leftarrow \var{Update}(c, \mathbf{E}_t)$\\
    }
    $\mathbf{E}_{t+1}^{r} \leftarrow (\mathbbm{1} \oslash \sum\limits_{c=1}^{C_t}{\mathbf{v}^{c})} \otimes \sum\limits_{c=1}^{C_t}{ \mathbf{E}_{t+1}^{r,c}}$ via \var{SecAgg}
  }
  \BlankLine
  \nonl \textbf{Client excutes} \var{Update$(c, \mathbf{E})$}\textbf{:}\\
  \For{\textup{each local epoch} $e = 1,2,...,E$}{
    \For{\textup{each batch} $\mathbf{b} = (\mathbf{h,r,t})$ \textup{of} $T^{c}$}{
        $\mathbf{E} \leftarrow \mathbf{E} - \eta \nabla \mathcal{L}, \text{where } \mathbf{E} := \{\mathbf{E}^{e,c}, \mathbf{E}^{r,c}\}$
    }
    \textup{Mask relation embedding:} $\mathbf{E}^{r,c} \leftarrow \mathbf{M}^{r,c} \otimes \mathbf{E}^{r,c}$
  }
  \Return{$\mathbf{E}^{r,c} \in \mathbf{E}, \mathbf{v}^c := \mathbf{M}^{r,c}$}
  \caption{\fedr{} Framework.}
  \label{alg:fkge}
\end{algorithm}
\vspace{-10pt}
\subsection{\fedr{}}

The overall procedure of \fedr{} framework is described in Algorithm \ref{alg:fkge}. Before aggregation works, the server acquires all IDs of the unique relations from local clients and maintains a relation table via Private Set Union (PSU), which computes the union of relations, without revealing anything else, for relation alignment \cite{kolesnikov2019scalable}. Hence, the server does not know the relations each client holds. The constructed relation table is then distributed to each client, and in each communication round, partial clients are selected to perform local training (see Appendix \ref{sec:local_update}) to update element embeddings $\mathbf{E}^c$ that will be masked by the masking indicator $\mathbf{M}^{r,c}$ and uploaded to the server later. Here $\mathbf{M}^{r,c}_i=1$ indicates the $i$-th entry in the relation table exists in client $c$. Considering that the server can retrive relations from each client by detecting if the embedidng is a vector of $\mathbf{0}$, we exploit Secure Aggregation technique (SecAgg, see Appendix \ref{sec:secagg}) in the aggregation phase as described in \textit{line 8} in Algorithm \ref{alg:fkge}, where $\oslash$ is element-wide division, $\otimes$ is element-wide multiplication, and $\mathbbm{1}$ is an all-one vector. The fundamental idea behind SecAgg is to mask the uploaded embeddings such that the server cannot obtain the actual ones from each client. However, the sum of masks can be canceled out, so we still have the correct aggregation results \citep{bonawitz2017practical}. Specifically, in \fedr{}, the server cannot access correct masking vectors $\mathbf{v}^{c}$ and embeddings $\mathbf{E}_{t+1}^{r,c}$ but only access the correct sum of them, namely, $\sum_{c=1}^{C_t}{\mathbf{v}^{c}}$ and $\sum_{c=1}^{C_t}{ \mathbf{E}_{t+1}^{r,c}}$, respectively. At the end of round $t$, the aggregated $\mathbf{E}_{t+1}^c$ will be sent back to each client $c \in C_t$ for next-round update.
\vspace{-5pt}
\section{Experiments}
We carry out several experiments to explore \fedr{}'s performance in link prediction, in which the tail $t$ is predicted given head $h$ and relation $r$.

\noindent\textbf{Datasets.} 
We evaluate our framework through experiments on three public datasets, FB15k-237, WN18RR \citep{dettmers2018convolutional} and a disease database -- DDB14 \citep{wang2021relational}. To build federated datasets, we randomly split triplets to each client without replacement. 
Note that, random split makes data heterogeneous among all the clients, and ensures fair comparison between FedE and FedR. 

\noindent\textbf{KGE Algorithms.} Four commonly-used KGE algorithms -- TransE \citep{bordes2013translating}, RotatE \citep{sun2019rotate}, DisMult \citep{yang2014embedding} and ComplEx \citep{trouillon2016complex} are utilized in the paper. We also implement federated NoGE \citep{Nguyen2022NoGE}, a GNN-based algorithm.

\begin{figure*}
     \centering
     \begin{subfigure}[b]{0.3\textwidth}
         \centering
         \includegraphics[width=\textwidth]{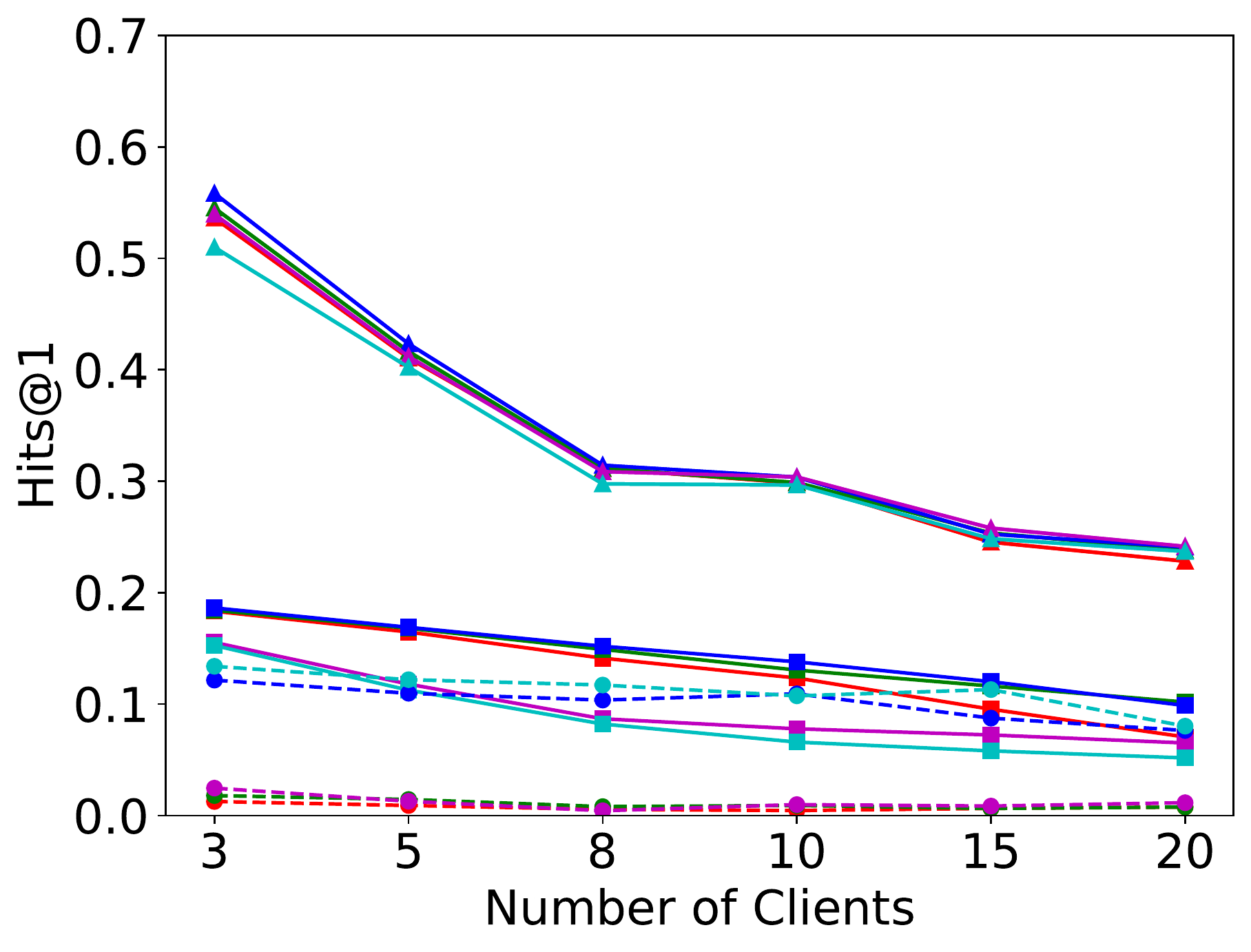}
     \end{subfigure}
     \hfill
     \begin{subfigure}[b]{0.3\textwidth}
         \centering
         \includegraphics[width=\textwidth]{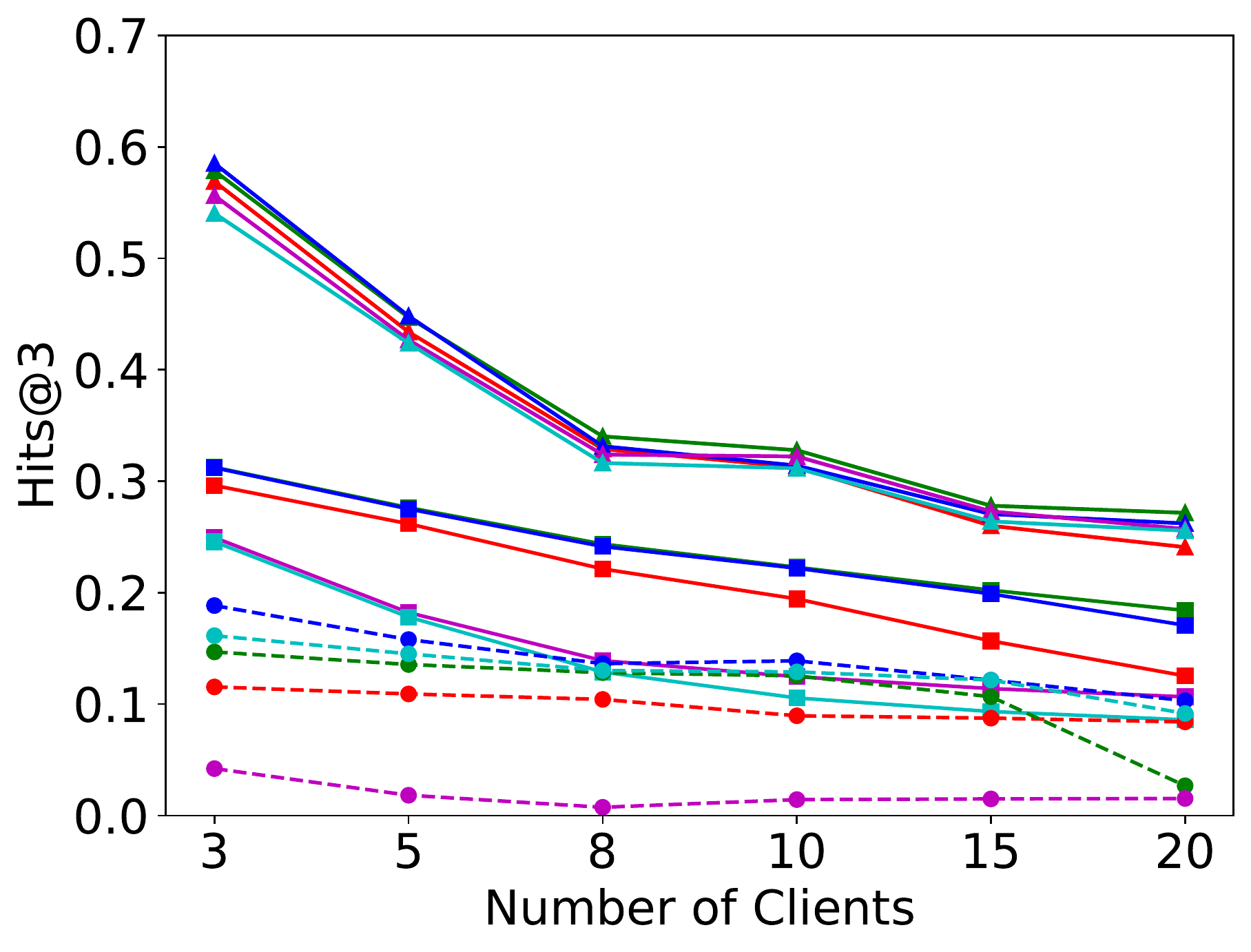}
     \end{subfigure}
     \hfill
     \begin{subfigure}[b]{0.3\textwidth}
         \centering
         \includegraphics[width=\textwidth]{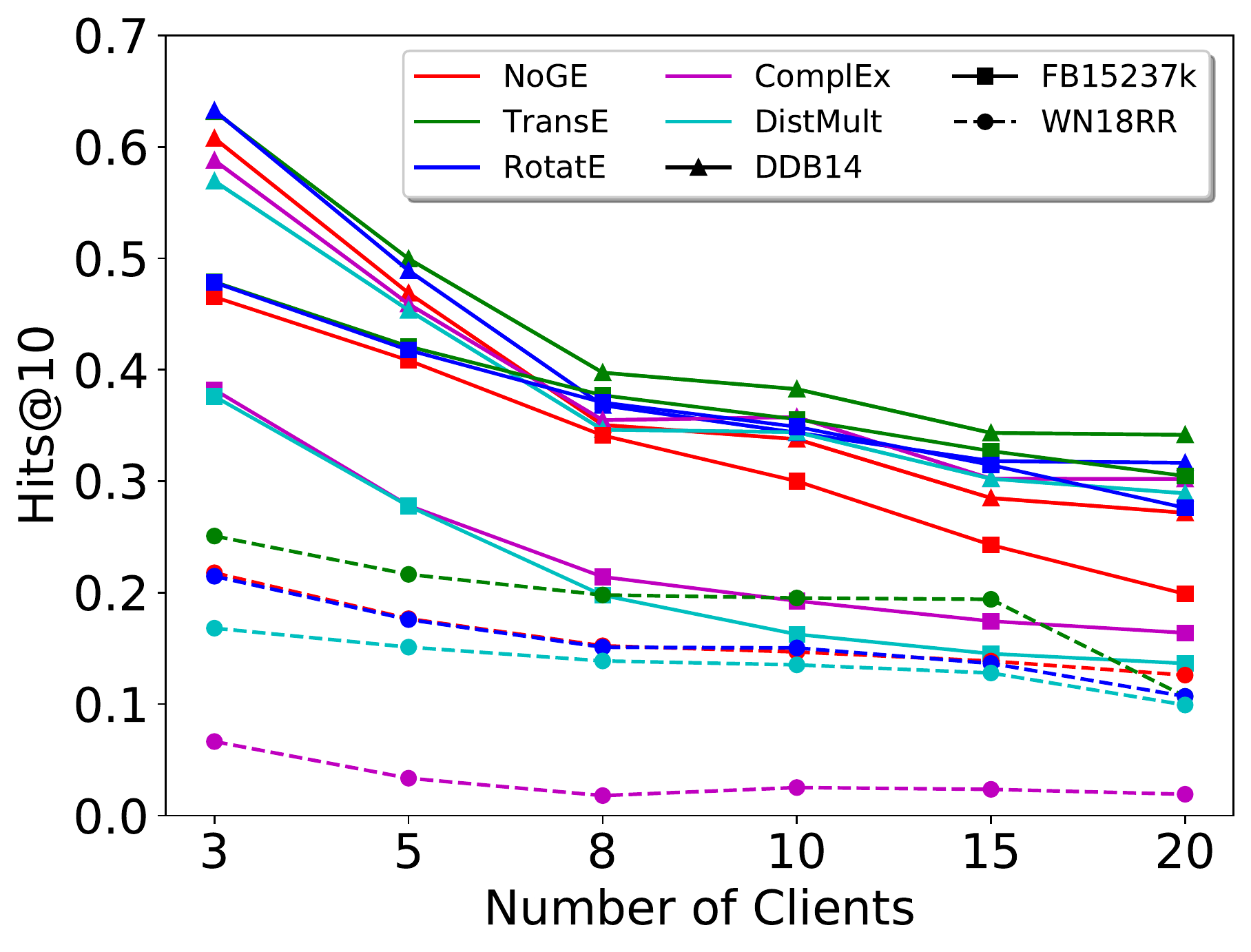}
     \end{subfigure}
        \caption{Experimental results of hit rates on three datasets.}
        \label{fig:hit_rate}
\end{figure*}

\subsection{Effectiveness Analysis} \label{sec:effect}
The commonly-used metric for link prediction, mean reciprocal rank (MRR), is exploited to evaluate \fedr{}'s performance. 
We take FedE and \var{Local}, where embeddings are trained only on each client's local KG, as the baselines. Table \ref{tab:effect} shows the link prediction results under settings of different number of clients $C$. We observe that \fedr{} comprehensively surpasses \var{Local} under all settings of the number of clients, which indicates that relation aggregation makes sense for learning better embeddings in FL. Take NoGE as an example, \fedr{} gains $29.64 \pm 0.037 \%$, $22.13 \pm 0.065 \%$, and $11.84 \pm 0.051 \%$ average improvement in MRR on three dataset. Compared with FedE, \fedr{} usually presents the better or similar results with the KGE models of DistMult and its extensive version ComplEx on all datasets. We also observe that both entity and relation aggregations succeed in beating \var{Local} setting but gain marginal improvement with DistMul and ComplEx on DDB14 and WN18RR datasets. Specially, 
KGE models fails to obtain reasonable results in federated with ComplEx. A potential reason could be that the averaging aggregation is not suitable for complex domains especially on the extremely unbalanced data (\textit{w.r.t} number of unique entities and relations in a KG).
Although FedE performs better than \fedr{} with TranE and RotatE, the absolute performance reductions between FedE and \fedr{} are mostly (13/16 = 81\%) within 0.03 in MRR on both DDB14 and FB15k-237, which illustrates that \fedr{} is still effective. The theoretical explanations behind these results \textit{w.r.t} data heterogeneity, and characteristics of FL and KGE models need further studies.

To further assess relation aggregation strategy, we compare performance of different KGE models regarding Hit Rates, which is shown in Figure \ref{fig:hit_rate}. Similar to MRR, Hit Rates drop with the increasing number of clients because of the more sparse knowledge distribution. 
All KGE models behave well and consistently on DDB14 dataset while there are large deviations of performance between each model on WN18RR and FB15k-237. This phenomenon is attributed to the biased local knowledge distribution, which is implicitly shown by the number of local entities. 

\subsection{Privacy Leakage Analysis} \label{sec:privacy}
Compared with entity aggregation, additional knowledge is required to perform reconstruction attack in \fedr{} because it is almost impossible to infer any entity or triplet from relation embeddings only. Therefore, we assume the server can access all entity embeddings without entity's IDs from clients. For simplicity, we let the server holds all information from C1, which is the same as the attack in Section \ref{sec:privacy_intro} (LR=100\%). The difference of adversary knowledge in FedE and \fedr{} is outlined in Table \ref{tab:adversary}. Besides, for fair comparison of FedE and \fedr{}, PSU and SecAgg are not considered.

\begin{table}[h]
\centering
\small
\begin{tabular}{ccccc}
\toprule
 & GEE & LEE & GRE & LRE \\ \midrule
FedE &\CheckmarkBold  &\CheckmarkBold  &\XSolidBrush  &\XSolidBrush  \\
FedR &\XSolidBrush  &\textcolor{red}{\CheckmarkBold}  &\CheckmarkBold  &\CheckmarkBold  \\ \bottomrule
\end{tabular}
\caption{Summary of adversary knowledge. ``G'' represents ``Global'', ``L'' represents ``Local''. ``EE'' and ``RE'' represent entity and relation embeddings, respectively.}
\label{tab:adversary}
\vspace{-5pt}
\end{table}

Table \ref{tab:privacy_fedr_other} presents the privacy leakage quantization in \fedr{} over three clients. The results shows that relation aggregation can protect both entity-level and graph-level privacy well even if providing additional local entity embeddings without considering encryption techniques. In addition, we observe that despite the relation embedding can be exploited directly in \fedr{} instead of inference, the privacy leakage rates in \fedr{} are still substantially lower than the ones in FedE. 
For example, according to Table \ref{tab:privacy_fb15k}, for C2, \fedr{} obtains relative reduction of 98.50\% and 99.52\% in ERR and TRR, respectively.
Note that once PSU and SecAgg are applied, \fedr{} can successfully defense against KG reconstruction attack and gain \textbf{NO} privacy leakage.

\begin{table}[h]
\centering
\setlength{\tabcolsep}{4.8pt}
\small
\begin{tabular}{lcccccc}
\toprule
\multirow{2}{*}{Dataset} & \multicolumn{2}{c}{FB15k-237} & \multicolumn{2}{c}{WN18RR} & \multicolumn{2}{c}{DDB14} \\ \cmidrule{2-7} 
         & ERR    & TRR    & ERR    & TRR   & ERR    & TRR   \\ \midrule
C2 \textbf{w/o} & 145.43 & 35.04 & 22.00 & 9.89 & 19.39 & 10.10  \\
C3 \textbf{w/o} & 129.77 & 22.01 & 18.44 & 9.23 & 8.87 & 5.05  \\ \hdashline
C2 \textbf{w} & \textbf{0} & \textbf{0} & \textbf{0} & \textbf{0} & \textbf{0} & \textbf{0}  \\
C3 \textbf{w} & \textbf{0} & \textbf{0} & \textbf{0} & \textbf{0} & \textbf{0} & \textbf{0}  \\ \bottomrule
\end{tabular}
\caption{Privacy leakage in \fedr{} with TransE ($\times 10^{-4}$).  \textbf{w} and \textbf{w/o} represent encryptions are applied or not.}
\label{tab:privacy_fedr_other}
\end{table}

\subsection{Communication Efficiency Analysis} \label{sec:comm}
\begin{figure}
    \centering
    \includegraphics[width=0.45\textwidth]{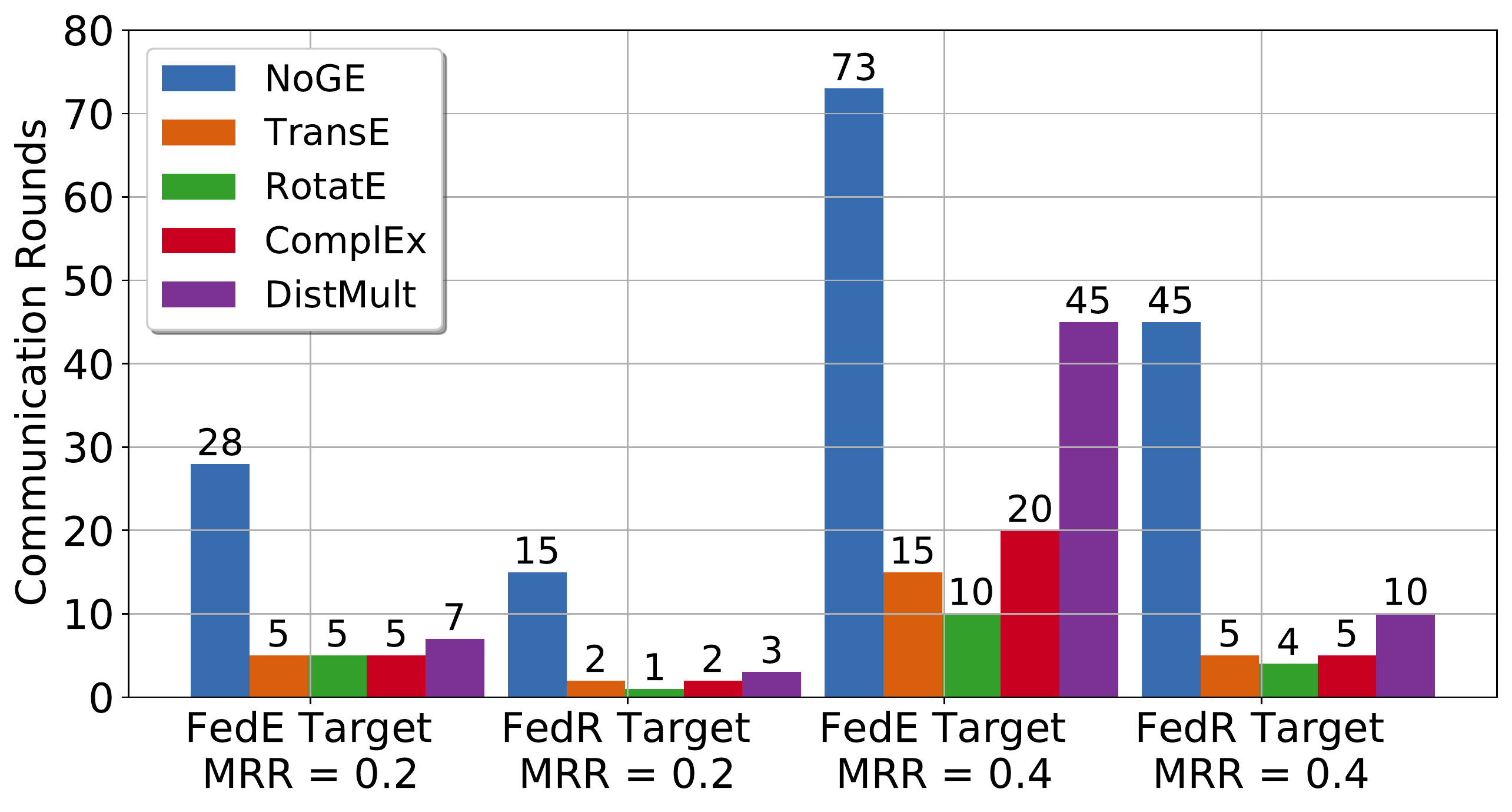}
    \vspace{-5pt}
    \caption{Number of communication rounds to reach a target MRR for FedE and \fedr{} with a fixed $C=5$.}
    \label{fig:comm}
    \vspace{-10pt}
\end{figure}

In this section, the product of data sizes and communication rounds is calculated to measure the communication cost. 
Considering the performance difference between \fedr{} and FedE, for fair comparison of communication efficiency, we count the rounds when the model reaches a pre-defined MRR target on the validation dataset. Specifically, we set two different MRR targets: 0.2 and 0.4. Since all models perform well on DDB14, we take the setting with $C=5$ on DDB14 as an example in this section. The required rounds for each model are depicted in Figure \ref{fig:comm}. We observe that \fedr{} reaches the target with much less rounds compared with FedE. For instance, \fedr{}-DistMult reaches the target MRR = 0.4 within 10 rounds while FedE uses 45 rounds. Also, according to statistics of federated datasets in Table \ref{tab:stat}, the average of the number of unique entities in FedE and unique relations in \fedr{} are 4462.2 and 12.8, respectively. We use the number of entities/relations to reflect data size, and by using relation aggregation, $99.89 \pm 0.029\%$ of cost is reduced in average for all clients when the target MRR is 0.2, while $99.90 \pm 0.042\%$ of cost is reduced in average when the target MRR is 0.4.
These results demonstrate that our proposed framework is more communication-efficient.

\subsection{Convergence Analysis}

\begin{figure}[h]
     \centering
     \begin{subfigure}[b]{0.4\textwidth}
         \centering
         \includegraphics[width=\textwidth]{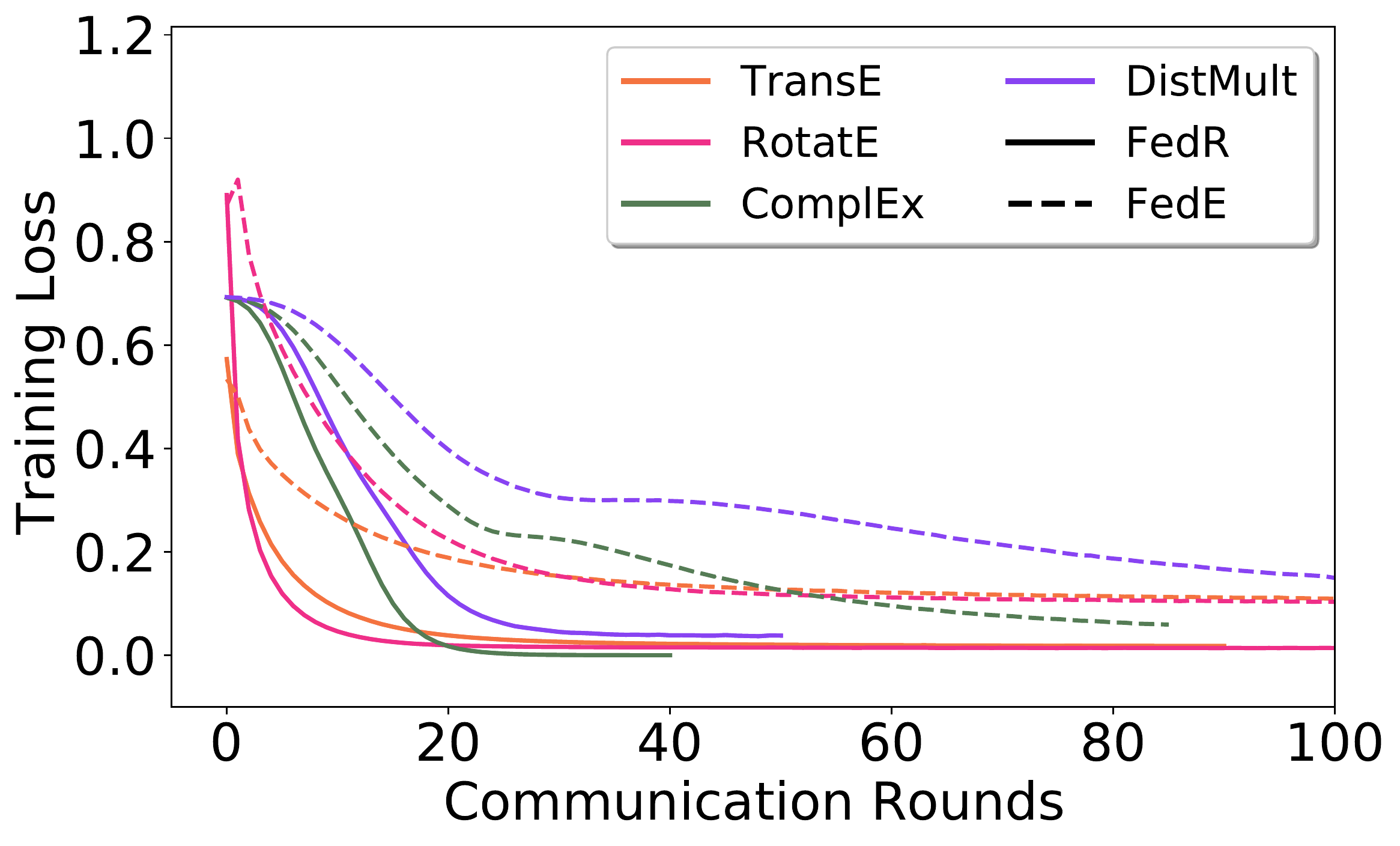}
         \caption{DDB14}
         \label{fig:loss_ddb}
     \end{subfigure}
     \hfill
     \begin{subfigure}[b]{0.4\textwidth}
         \centering
         \includegraphics[width=\textwidth]{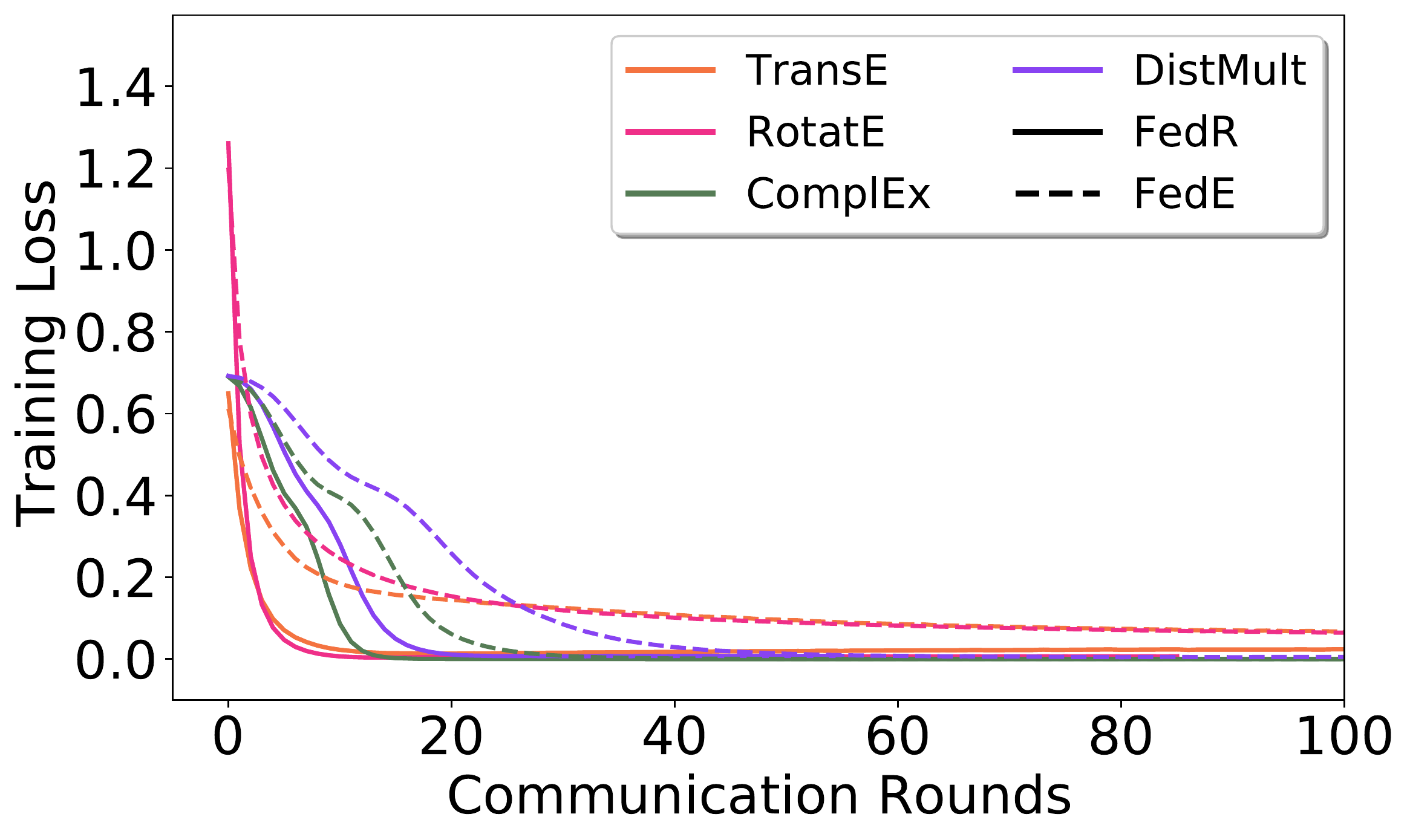}
         \caption{WN18RR}
         \label{fig:loss_wn18}
     \end{subfigure}
     \hfill
     \begin{subfigure}[b]{0.4\textwidth}
         \centering
         \includegraphics[width=\textwidth]{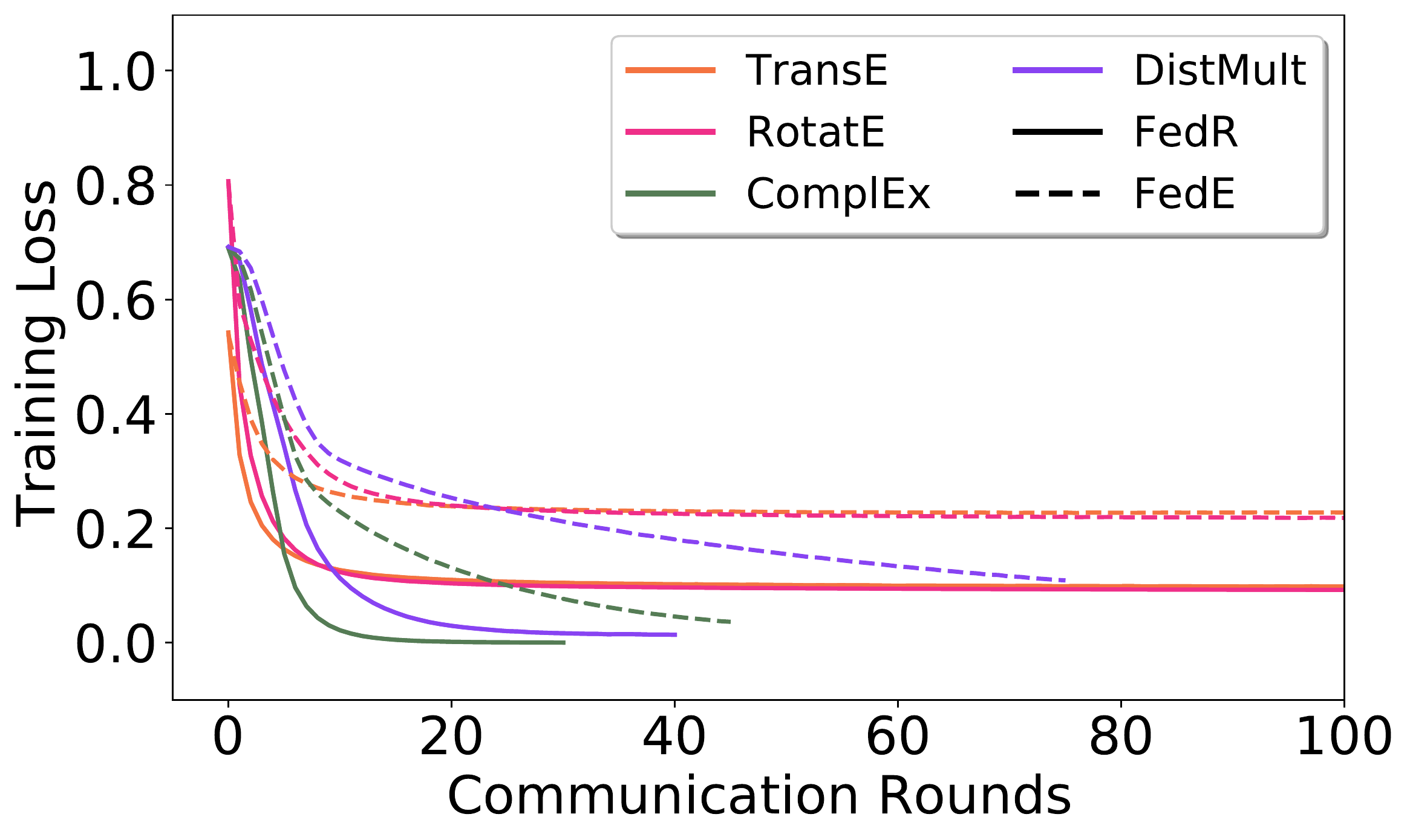}
         \caption{FB15k-237}
         \label{fig:loss_fb15k}
     \end{subfigure}
        \caption{Training  loss  versus  communication ($C= 5$).}
        \vspace{-10 pt}
        \label{fig:loss}
\end{figure}

The convergence curves considering four KGE models and three dataset are shown in Figure \ref{fig:loss}. The solid and dashed lines represent curves \textit{w.r.t} \fedr{} and FedE, respectively. We do not show the curves of NoGE because the aggregated embeddings does not influence local training. We observe that \fedr{} usually converge faster than FedE. Some lines are incomplete over communication rounds because early-stop technique in terms of validation MRR is used in the experiments. 

\section{Conclusion and Future Work}


In this paper, we conduct the first empirical quantization of privacy leakage to federated learning on knowledge graphs, which reveals that recent work, FedE, is susceptible to reconstruction attack based on shared element-embedding pairs when there are dishonest server and clients. Then we propose \fedr{}, a privacy-preserving FL framework on KGs with relation embedding aggregation that defenses against reconstruction attack effectively. Experimental results show that \fedr{} outperforms FedE w.r.t data privacy and communication efficiency and also maintains similar utility.

In real-world applications, different organizations may use different KGE models, which may influence overall performance by embedding aggregation, how to design an effective FL framework in this case and how to perform KG reconstruction attack/defense are our future research directions.

\section{Limitations}
Both \fedr{} and FedE are sensitive to data distribution. For example,  if we build subgraphs in terms of relations, \fedr{} may not effective because of less overlapping relations among clients. It is still an open question that how to develop an FL architecture over arbitrarily non-iid KGs.

\bibliography{anthology,custom}
\bibliographystyle{acl_natbib}

\appendix




\section{Knowledge Graph Reconstruction}
\label{sec:kg_attack}

We summarize the knowledge graph reconstruction attack in Algorithm \ref{alg:kgr}. Note that in the algorithm, i) and ii) refer to different operations, and only one will be performed in FedE or \fedr{}.

\begin{algorithm}
  \nonl \textbf{Adversary knowledge:} Local entity embeddings -- $\mathbf{LEE}$, \textcolor{red}{local relation embeddings -- $\mathbf{LRE}$}, element-embedding paris from a client -- $\mathbf{EEP}$, type of the used KGE model. \\
  \BlankLine
  \nonl \textbf{Entity reconstruction:} \\
  \For{\textup{entity embedding} $\hat{e} \in \mathbf{LEE}$}{
    \For{\textup{entity-embedding} $(E, e) \in \mathbf{EEP}$}{
    \textup{Calculate similarity between $e$ and $\hat{e}$}\\
    \textup{Update the inferred entity} $\hat{E} = E$ with the greatest similarity score\\}
    }
  \Return the reconstructed entity set {$\{\hat{E}\}$}
  
  \BlankLine
  \nonl \textbf{Triple reconstruction:} \\
  \nonl \textcolor{blue}{only one of i) and ii) will be implemented}\\
  i) \For{\textup{entity embeddings} $(\hat{h}, \hat{t}) \in \mathbf{LEE}$}{
    \textup{Calculate relation embedding} $\hat{r}$ based on the scoring function of used KGE model, e.g. $\hat{r} = \hat{t} - \hat{h}$ with TransE \\
    \For{\textup{relation-embedding}$(R,r) \in \mathbf{EEP}$}{
        Calculate similarity between $r$ and $\hat{r}$ \\
        Update the inferred relation $\hat{R} = R$ with the greatest similarity score \\}
    }
    \Return the reconstructed relation set $\{\hat{R}\}$\\
    
  \BlankLine
  \textcolor{red}{ii)} \For{\textup{\textcolor{red}{relation embedding}} \textcolor{red}{$\hat{r} \in \mathbf{LRE}$}}{
    \For{\textcolor{red}{\textup{relation-embedding}$(R,r) \in \mathbf{EEP}$}}{
        \textcolor{red}{Calculate similarity between $r$ and $\hat{r}$} \\
        \textcolor{red}{Update the inferred relation $\hat{R} = R$ with the greatest similarity score} \\}
    }
  \Return \textcolor{red}{the reconstructed relation set $\{\hat{R}\}$}\\

  \BlankLine
  Utilize $\{\hat{E}\}$ and $\{\hat{R}\}$ to reconstruct triples.
  
  \caption{Knowledge graph reconstruction including attack in \fede{}/\textcolor{red}{\fedr{}}.}
  \label{alg:kgr}
\end{algorithm}

\section{Implementation Details}
\label{sec:impelment}

For TransE, RotatE, DistMult, and ComplEx, we follow the same setting as FedE \citep{chen2021fede}. Specifically, the number of negative sampling, margin $\gamma$ and the negative sampling temperature $\alpha$ are set as 256, 10 and 1, respectively. Note that, we adopt a more conservative strategy for embedding aggregation where local non-existent entities will not be taken as negative samples compared to FedE. For NoGE, we use GCN \citep{kipf2016semi} as encoder and QuatE \citep{zhang2019quaternion} as decoder. Once local training is done in a communciation round, the embeddings are aggregated and the triplet is scored by the decoder. The hidden size of 1 hidden layer in NoGE is 128. 

If not specified, the local update epoch is 3, the embedding dimension of entities and relation is 128. Early stopping is utilized in experiments. The patience, namely the number of epochs with no improvement in MRR on validation data after which training will be stopped, is set as 5. We use Adam with learning rate $0.001$ for local model update. All models are trained using one Nvidia 2080 GPU with 300 communication rounds at maximum.

\begin{table}[]
\centering
\small
\begin{tabular}{cccccc}
\toprule
Dataset & \#C & \#Entity & \#Relation \\ \midrule
\multirow{4}{*}{DDB14}
 & 5  &4462.20$_{\pm 1049.60}$  &12.80$_{\pm 0.84}$\\  
 & 10 &3182.60$_{\pm 668.89}$  &12.60$_{\pm 0.70}$\\ 
 & 15 &2533.86$_{\pm 493.47}$  &12.50$_{\pm 0.74}$\\  
 & 20 &2115.59$_{\pm 385.56}$  &12.35$_{\pm 0.75}$\\ \midrule 
\multirow{4}{*}{WN18RR}
 & 5  &21293.20$_{\pm 63.11}$  &11.00$_{\pm 0.00}$ \\
 & 10 &13112.20$_{\pm 46.70}$  &11.00$_{\pm 0.00}$ \\
 & 15 &9537.33$_{\pm 45.45}$  &11.00$_{\pm 0.00}$  \\ 
 & 20 &7501.65$_{\pm 31.72}$  &11.00$_{\pm 0.00}$ \\  \midrule
\multirow{4}{*}{FB15k-237}
 & 5  &13359.20$_{\pm 27.36}$  &237.00$_{\pm 0.00}$ \\
 & 10 &11913.00$_{\pm 31.56}$  &237.00$_{\pm 0.00}$ \\
 & 15 &10705.87$_{\pm 36.93}$  &236.87$_{\pm 0.35}$  \\ 
 & 20 &9705.95$_{\pm 44.10}$  &236.80$_{\pm 0.41}$ \\ \bottomrule
\end{tabular}
\caption{Statistics of federated datasets. 
The subscripts denote standard deviation. \# denotes ``number of''.}
\label{tab:stat}
\end{table}

\subsection{Statistics of Datasets}
To build federated datasets, we randomly split triples to each client without replacement, then divide the local triples into the train, valid, and test sets with a ratio of 80/10/10. The statistics of datasets after split is
described in Table \ref{tab:stat}. 

\subsection{Client Update} \label{sec:local_update}
The client update, or loca knowledge graph embedding update, corresponds to \var{Update$(c, \mathbf{E})$} in Algorithm \ref{alg:fkge} starting from \textit{line 9}, which learns both embeddings of entities and relations. 

For a triplet $(h,r,t)$ in client $c$, we adopt the self-adversarial nagative sampling \citep{sun2019rotate} for effectively optimizing non-GNN KGE models:
\begin{equation*} 
\begin{split}
    &\mathcal{L}(h,r,t) = -\log \sigma (\gamma - f_{r}(\mathbf{h,t})) \\
    &- \sum\limits_{i=1}^n p(h, r, t_i') \log \sigma (f_{r}(\mathbf{h,} \mathbf{t}_i^{\prime}) - \gamma),
\end{split}
\end{equation*}
where $\gamma$ is a predefined margin, $\sigma$ is the sigmoid function, $f$ is the scoring function that varies as shown in Table \ref{tab:score_func}, and $(\mathbf{h}, \mathbf{r}, \mathbf{t}_i^{\prime})$ is the $i$-th negative triplet, which can be sampled from the following distribution:
\begin{equation*}
    p(h, r, t_{j}^{\prime} | \{(h_{i}, r_{i}, t_{i})\})=\frac{\exp \alpha f_{r}(\mathbf{h,} \mathbf{t}_i^{\prime})}{\sum_{i} \exp \alpha f_{r}(\mathbf{h,} \mathbf{t}_i^{\prime})}
\end{equation*}
where $\alpha$ is the temperature of sampling. There would be $E$ epoches of traning on the client at a round to update local-view embeddings $\mathbf{E}$ including entity and relation embeddings, but only local relation embeddings $\{\mathbf{E}^{r,c}\}$ will be sent to server.

For NoGE, we follow its plain design by minimizing the binary cross-entryopy loss function:
\begin{equation*}
    \begin{split}
        \mathcal{L}&=-\sum_{(h, r, t)} (l_{(h, r, t)} \log \left(\var{sigmoid}(f(\mathbf{h,r,t}))\right) \\
        &+ \left(1-l_{(h, r, t)}\right) \log \left(1-\var{sigmoid}(f(\mathbf{h,r,t})\right)) \\
    \end{split}
\end{equation*}
\begin{equation*}
    \text { in which, } l_{(h, r, t)}= \begin{cases}1 & \text { for }(h, r, t) \in G \\
0 & \text { for }(h, r, t) \in G^{\prime}\end{cases}
\end{equation*}
where $G$ and $G^{\prime}$ are collections of valid and invalid triplets, respectively.

\subsection{Scoring Function}
\label{sec:score_func}

\begin{table}[htbp]
\centering
\small
\begin{tabular}{cc}
\toprule
Model & Scoring Function \\ \midrule
TransE & $-\|\mathbf{h}+\mathbf{r}-\mathbf{t}\|$ \\
RotatE & $-\|\mathbf{h} \circ \mathbf{r}-\mathbf{t}\|$ \\
DistMult & $\mathbf{h}^{\top} \operatorname{diag}(\mathbf{r}) \mathbf{t}$ \\
ComplEx & $\operatorname{Re}\left(\mathbf{h}^{\top} \operatorname{diag}(\mathbf{r}) \overline{\mathbf{t}}\right)$ \\
NoGE & $\left\langle a_{h}^{\prime}, a_{t}\right\rangle+\left\langle b_{h}^{\prime}, b_{t}\right\rangle+\left\langle c_{h}^{\prime}, c_{t}\right\rangle+\left\langle d_{h}^{\prime}, d_{t}\right\rangle$ \\
KB-GAT & $\left(\|_{m=1}^{\Omega} \operatorname{ReLU}\left(\left[\vec{h}_{i}, \vec{g}_{k}, \vec{h}_{j}\right] * \omega^{m}\right)\right) \cdot \mathbf{W}$ \\
\bottomrule
\end{tabular}
\caption{A list of scoring functions for KGE models implemented in this paper. The scoring function used in NoGE comes from QuatE \cite{zhang2019quaternion}.}
\label{tab:score_func}
\end{table}

\section{Secure Aggregation in \fedr{}} \label{sec:secagg}
In this section, we illustrate how SecAgg works in \fedr{} through a simple exmaple including three clients with two relations. Mathematically, we assume the distribution of relation embeddings as $\mathbf{R}_1 = \{r_1\}, \mathbf{R}_2 = \{r_2\}$ and $\mathbf{R}_3 = \{r_1\}$, respectively. After PSU, the server will obtain a set of relations $\mathbf{R} = \{r_1, r_2\}$. Besides, we denote the corresponding masking vectors as $\mathbf{M}_1 = (1, 0), \mathbf{M}_2 = (0, 1) \textup{ and } \mathbf{M}_3 = (1, 0)$.

In one communication round, once all clients complete local training and prepare for the aggregation phase, via Diffie-Hellman secret sharing \cite{bonawitz2017practical}, each client $u$ generates $s_{u,v}$ randomly for every other client, and they agree on the large prime number $l$. Then each party $u$ compute the masked value $t_u$ for its secret vector $s_u$, where $s_u := \{\mathbf{R}_u, \mathbf{M}_u\}$, shown as below:
\begin{equation*}
    t_u = s_u + \sum_{u<v} s_{u,v} - \sum_{u>v} s_{v,u} \;\;\; (\text{mod } l),
\end{equation*}
where $s_{u,v} = s_{v,u}$ for a specific condition, e.g. $s_{1,2}=s_{2,1}$. Therefore, each client holds its masked matrix as follows:
\begin{equation*}
\begin{split}
    &t_1 = s_1 + s_{1,2} + s_{1,3} \;\;\; (\text{mod } l), \\
    &t_2 = s_2 + s_{2,3} - s_{2,1} \;\;\; (\text{mod } l), \\
    &t_3 = s_3 - s_{3,1} - s_{3,2} \;\;\; (\text{mod } l), \\
\end{split}
\end{equation*}
Next, these masked matrices are uploaded to the server. Now the server cannot obtain the actual information from clietns but could extract the correct aggregated value via: 
\begin{equation*}
\begin{split}
\mathbf{z} &= \sum_{u=1}^3 t_u \\
&= \sum_{u,v=1}^3 (s_u + \sum_{u<v} s_{u,v} - \sum_{u>v} s_{v,u}) \\
&= \sum_{u=1}^3 s_u \;\;\; (\text{mod } l)
\end{split}
\end{equation*}


\section{Additional Results}
\label{sec:extensive}


In this section, we introduce additional experimental results of KB-GAT in a federated manner for link prediction.

\subsection{Experiment result with KB-GAT}
Since the aggregated information is not exploited in the local training in NoGE, we also implement KB-GAT \cite{nathani2019learning}, the other GNN model but it can take advantages of both graph structure learning and global-view information aggregation. However, Fed-KB-GAT is memory-consuming. For KB-GAT, we use GAT \citep{velivckovic2018graph} as encoder and ConvKB \citep{nguyen2018novel} as decoder. Although the input to KB-GAT is the triple embedding, this model update neural network weights to obtain the final entity and relation embeddings. In each communication, we let the aggregated embeddings be the new input to KB-GAT, we find using small local epoches lead to bad performance because the model is not fully trained to produce high-quality embeddings. Therefore, we set local epoch of GAT layers as 500, while local epoch of convlutional layers as 150. Embedding size is 50 instead of 128 like others since we suffers memory problem using this model.

We conduct KB-GAT with both entity aggregation and relation aggregation on DDB14 with $C=3$ as shown in Table \ref{tab:kb-gat}. Due to the good performance of RotatE, we also compare KB-GAT with RotatE. Hit@N is also utilized in the evaluation. From the table, KB-GAT beats RotatE in regard of all evaluation metrics in both FedE and FedR setting. However, how to implement federated KB-GAT in a memory-efficient way is still an open problem.

\begin{table}[]
\centering
\setlength{\tabcolsep}{4.0pt}
\small
\begin{tabular}{cccccc}
\toprule
Model & Setting & MRR & Hit@1 & Hit@3 & Hit@10 \\ \midrule
\multirow{3}{*}{RotatE}
 & \var{Local} &0.5347  &0.5311  &0.5459  &0.5912  \\
 & FedE &0.6087  &0.5070  &0.6774  &0.7916  \\
 & \fedr{} &0.5834  &0.5583  &0.5852  &0.6326 \\ \midrule
\multirow{3}{*}{KB-GAT}
 & \var{Local} &0.4467  &0.4369  &0.4620  &0.4755  \\
 & FedE &\textbf{0.5622}  &\textbf{0.5471}  &\textbf{0.5634}  & \textbf{0.5887}  \\
 & \fedr{} &\underline{0.5034}  &\underline{0.4861}  &\underline{0.5301}  &\underline{0.5644}  \\ \bottomrule
\end{tabular}
\caption{\small{Extensive experimental resutls on DDB14 with $C=3$. \textbf{Bold} number denotes the best result in FedE and \underline{underline} number denotes the best result in \fedr{}}.}
\label{tab:kb-gat}
\end{table}

\end{document}